\documentclass[a4paper]{article}

\usepackage{INTERSPEECH2016}

\usepackage{graphicx}
\usepackage{amssymb,amsmath,bm}
\usepackage{textcomp}

\ninept

\title{Attention-Based Recurrent Neural Network Models for Joint Intent Detection and Slot Filling}

\makeatletter
\def\name#1{\gdef\@name{#1\\}}
\makeatother \name{{\em Bing Liu$^1$, Ian Lane$^1$$^,$$^2$}}

\address{
  $^1$Electrical and Computer Engineering, Carnegie Mellon University \\
  $^2$Language Technologies Institute, Carnegie Mellon University \\
  {\small \tt liubing@cmu.edu, lane@cmu.edu}
}

\begin{document}

  \maketitle
  \begin{abstract}
    Attention-based encoder-decoder neural network models have recently shown promising results in machine translation and speech recognition. In this work, we propose an attention-based neural network model for joint intent detection and slot filling, both of which are critical steps for many speech understanding and dialog systems. Unlike in machine translation and speech recognition, alignment is explicit in slot filling. We explore different strategies in incorporating this alignment information to the encoder-decoder framework. Learning from the attention mechanism in encoder-decoder model, we further propose introducing attention to the alignment-based RNN models. Such attentions provide additional information to the intent classification and slot label prediction. Our independent task models achieve state-of-the-art intent detection error rate and slot filling F1 score on the benchmark ATIS task. Our joint training model further obtains 0.56\% absolute (23.8\% relative) error reduction on intent detection and 0.23\% absolute gain on slot filling over the independent task models.
  \end{abstract}
  \noindent{\bf Index Terms}: Spoken Language Understanding, Slot Filling, Intent Detection, Recurrent Neural Networks, Attention Model

\section{Introduction}

    Spoken language understanding (SLU) system is a critical component in spoken dialogue systems. SLU system typically involves identifying speaker's intent and extracting semantic constituents from the natural language query, two tasks that are often referred to as intent detection and slot filling. 
    
    Intent detection and slot filling are usually processed separately. Intent detection can be treated as a semantic utterance classification problem, and popular classifiers like support vector machines (SVMs) \cite{haffner2003optimizing} and deep neural network methods \cite{sarikaya2011deep} can be applied. Slot filling can be treated as a sequence labeling task. Popular approaches to solving sequence labeling problems include maximum entropy Markov models (MEMMs) \cite{mccallum2000maximum}, conditional random fields (CRFs) \cite{raymond2007generative}, and recurrent neural networks (RNNs) \cite{yao2014spoken, mesnil2015using, liu2015recurrent}. Joint model for intent detection and slot filling has also been proposed in literature \cite{guo2014joint, xu2013convolutional}. Such joint model simplifies the SLU system, as only one model needs to be trained and fine-tuned for the two tasks. 
    
    Recently, encoder-decoder neural network models have been successfully applied in many sequence learning problems such as machine translation \cite{sutskever2014sequence} and speech recognition \cite{chan2015listen}. The main idea behind the encoder-decoder model is to encode input sequence into a dense vector, and then use this vector to generate corresponding output sequence. The attention mechanism introduced in \cite{bahdanau2014neural} enables the encoder-decoder architecture to learn to align and decode simultaneously. 
    
    In this work, we investigate how an SLU model can benefit from the strong modeling capacity of the sequence models. Attention-based encoder-decoder model is capable of mapping sequences that are of different lengths when no alignment information is given. In slot filling, however, alignment is explicit, and thus alignment-based RNN models typically work well. We would like to investigate the combination of the attention-based and alignment-based methods. Specifically, we want to explore how the alignment information in slot filling can be best utilized in the encoder-decoder models, and on the other hand, whether the alignment-based RNN slot filling models can be further improved with the attention mechanism that introduced from the encoder-decoder architecture. Moreover, we want to investigate how slot filling and intent detection can be jointly modeled under such schemes.

    The remainder of the paper is organized as follows. In section 2, we introduce the background on using RNN for slot filling and using encoder-decoder models for sequence learning. In section 3, we describe two approaches for jointly modeling intent and slot filling. Section 4 discusses the experiment setup and results on ATIS benchmarking task. Section 5 concludes the work.

\section{Background}

    \subsection{RNN for Slot Filling}
    
        Slot filling can be treated as a sequence labeling problem, where we have training examples of $\left \{ (\mathbf{x}^{(n)}, \mathbf{y}^{(n)}): n=1,...,N \right \}$ and we want to learn a function $f:\mathcal{X} \rightarrow \mathcal{Y}$ that maps an input sequence $\mathbf{x}$ to the corresponding label sequence $\mathbf{y}$. In slot filling, the input sequence and label sequence are of the same length, and thus there is explicit alignment.
        
            \begin{figure}[h]
                \centering
                \includegraphics[width=220pt]{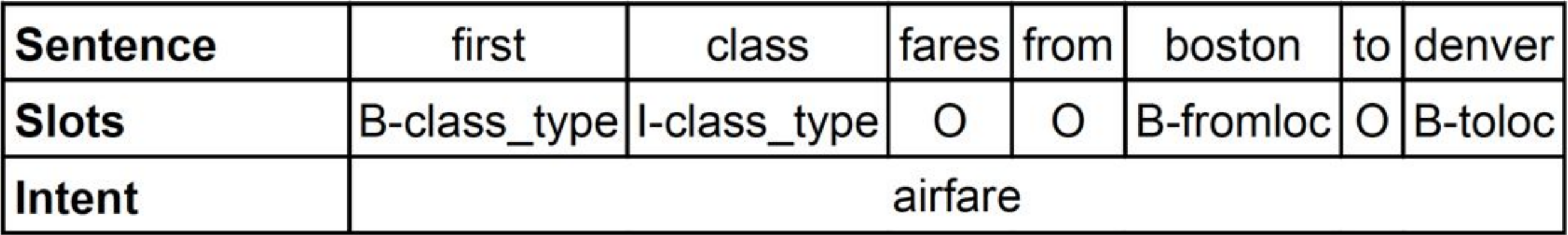}
                \caption{{\it ATIS corpus sample with intent and slot annotation. }}
                \label{fig:slot_filling_example.pdf}
            \end{figure}
        
        RNNs have been widely used in many sequence modeling problems \cite{mesnil2015using, mikolov2011extensions}. At each time step of slot filling, RNN reads a word as input and predicts its corresponding slot label considering all available information from the input and the emitted output sequences. The model is trained to find the best parameter set $\theta$ that maximizes the likelihood:
        \begin{equation}
            \arg\max_\theta \prod_{t=1}^{T}P(y_{t}|y_{1}^{t-1}, \mathbf{x}; \theta) \\
        \end{equation}
        where $\mathbf{x}$ represents the input word sequence, $y_{1}^{t-1}$ represents the output label sequence prior to time step $t$. During inference, we want to find the best label sequence $\mathbf{y}$ given an input sequence $\mathbf{x}$ such that:
        \begin{equation}
            \hat{\mathbf{y}} = \operatorname{\arg\max_\mathbf{y}} P(\mathbf{y}|\mathbf{x}) \\
        \end{equation}
    
    \subsection{RNN Encoder-Decoder}
    
        The RNN encoder-decoder framework is firstly introduced in \cite{sutskever2014sequence} and \cite{cho2014learning}. The encoder and decoder are two separate RNNs. The encoder reads a sequence of input $(x_{1}, ..., x_{T})$ to a vector $c$. This vector encodes information of the whole source sequence, and is used in decoder to generate the target output sequence. The decoder defines the probability of the output sequence as:
        \begin{equation}
            P(\mathbf{y}) = \prod_{t=1}^{T}P(y_{t}|y_{1}^{t-1}, c) \\
        \end{equation}
        where $y_{1}^{t-1}$ represents the predicted output sequence prior to time step $t$.
        Comparing to an RNN model for sequence labeling, the RNN encoder-decoder model is capable of mapping sequence to sequence with different lengths. There is no explicit alignment between source and target sequences. The attention mechanism later introduced in \cite{bahdanau2014neural} enables the encoder-decoder model to learn a soft alignment and to decode at the same time.

\section{Proposed Methods}
    In this section, we first describe our approach on integrating alignment information to the encoder-decoder architecture for slot filling and intent detection. Following that, we describe the proposed method on introducing attention mechanism from the encoder-decoder architecture to the alignment-based RNN models. 

    \subsection{Encoder-Decoder Model with Aligned Inputs}
    
        \begin{figure}[t]
            \centering
            \includegraphics[width=\linewidth]{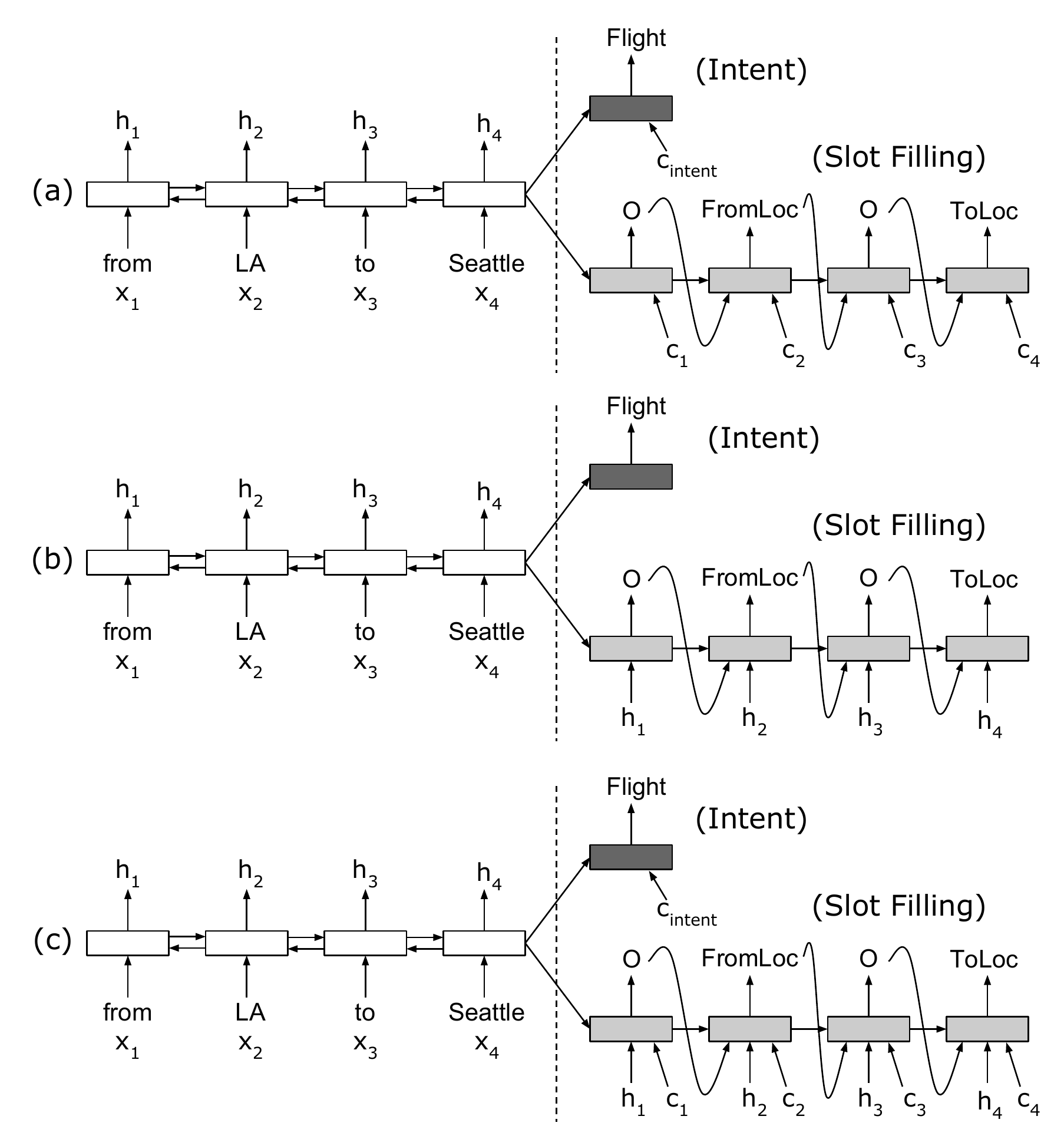}
            \caption{{\it  Encoder-decoder model for joint intent detection and slot filling. (a) with no aligned inputs. (b) with aligned inputs. (c) with aligned inputs and attention. Encoder is a bidirectional RNN. The last hidden state of the backward encoder RNN is used to initialize the decoder RNN state.}}
            \label{fig:alignment_encoder-decoder}
        \end{figure}
        
        The encoder-decoder model for joint intent detection and slot filling is illustrated in Figure \ref{fig:alignment_encoder-decoder}. On encoder side, we use a bidirectional RNN. Bidirectional RNN has been successfully applied in speech recognition \cite{graves2013hybrid} and spoken language understanding \cite{mesnil2015using}. We use LSTM \cite{hochreiter1997long} as the basic recurrent network unit for its ability to better model long-term dependencies comparing to simple RNN.
        
        In slot filling, we want to map a word sequence $\mathbf{x} = (x_{1}, ..., x_{T})$ to its corresponding slot label sequence $\mathbf{y} = (y_{1}, ..., y_{T})$. The bidirectional RNN encoder reads the source word sequence forward and backward. The forward RNN reads the word sequence in its original order and generates a hidden state $fh_{i}$ at each time step. Similarly, the backward RNN reads the word sequence in its reverse order and generate a sequence of hidden states $(bh_{T}, ..., bh_{1})$. The final encoder hidden state $h_{i}$ at each time step $i$ is a concatenation of the forward state $fh_{i}$ and backward state $bh_{i}$, i.e. $h_{i} = [fh_{i}, bh_{i}]$.
        
        The last state of the forward and backward encoder RNN carries information of the entire source sequence. We use the last state of the backward encoder RNN to compute the initial decoder hidden state following the approach in \cite{bahdanau2014neural}. The decoder is a unidirectional RNN. Again, we use an LSTM cell as the basic RNN unit. At each decoding step $i$, the decoder state $s_{i}$ is calculated as a function of the previous decoder state $s_{i-1}$, the previous emitted label $y_{i-1}$, the aligned encoder hidden state $h_{i}$, and the context vector $c_{i}$:
        \begin{equation}
            s_{i} = f(s_{i-1}, y_{i-1}, h_{i}, c_{i})
        \end{equation}
        where the context vector $c_{i}$ is computed as a weighted sum of the encoder states $\mathbf{h} = (h_{1}, ..., h_{T})$ \cite{bahdanau2014neural}:
        \begin{equation}
            c_{i} = \sum_{j=1}^{T}\alpha_{i,j}h_{j}
        \end{equation}        
        and
        \begin{equation}
        \begin{split}
            \alpha_{i,j} = \frac{\operatorname{exp}(e_{i, j})}{\sum_{k=1}^T\operatorname{exp}(e_{i, k})} \\ 
            e_{i, k} = g(s_{i-1}, h_{k})
        \end{split}
        \end{equation} 
        
        $g$ a feed-forward neural network. At each decoding step, the explicit aligned input is the encoder state $h_{i}$. The context vector $c_{i}$ provides additional information to the decoder and can be seen as a continuous bag of weighted features $(h_{1}, ..., h_{T})$.
        
        For joint modeling of intent detection and slot filling, we add an additional decoder for intent detection (or intent classification) task that shares the same encoder with slot filling decoder. During model training, costs from both decoders are back-propagated to the encoder. The intent decoder generates only one single output which is the intent class distribution of the sentence, and thus alignment is not required. The intent decoder state is a function of the shared initial decoder state $s_{0}$, which encodes information of the entire source sequence, and the context vector $c_{intent}$, which indicates part of the source sequence that the intent decoder pays attention to.

    \subsection{Attention-Based RNN Model}
    
        \begin{figure}[t]
            \centering
            \includegraphics[width=220pt]{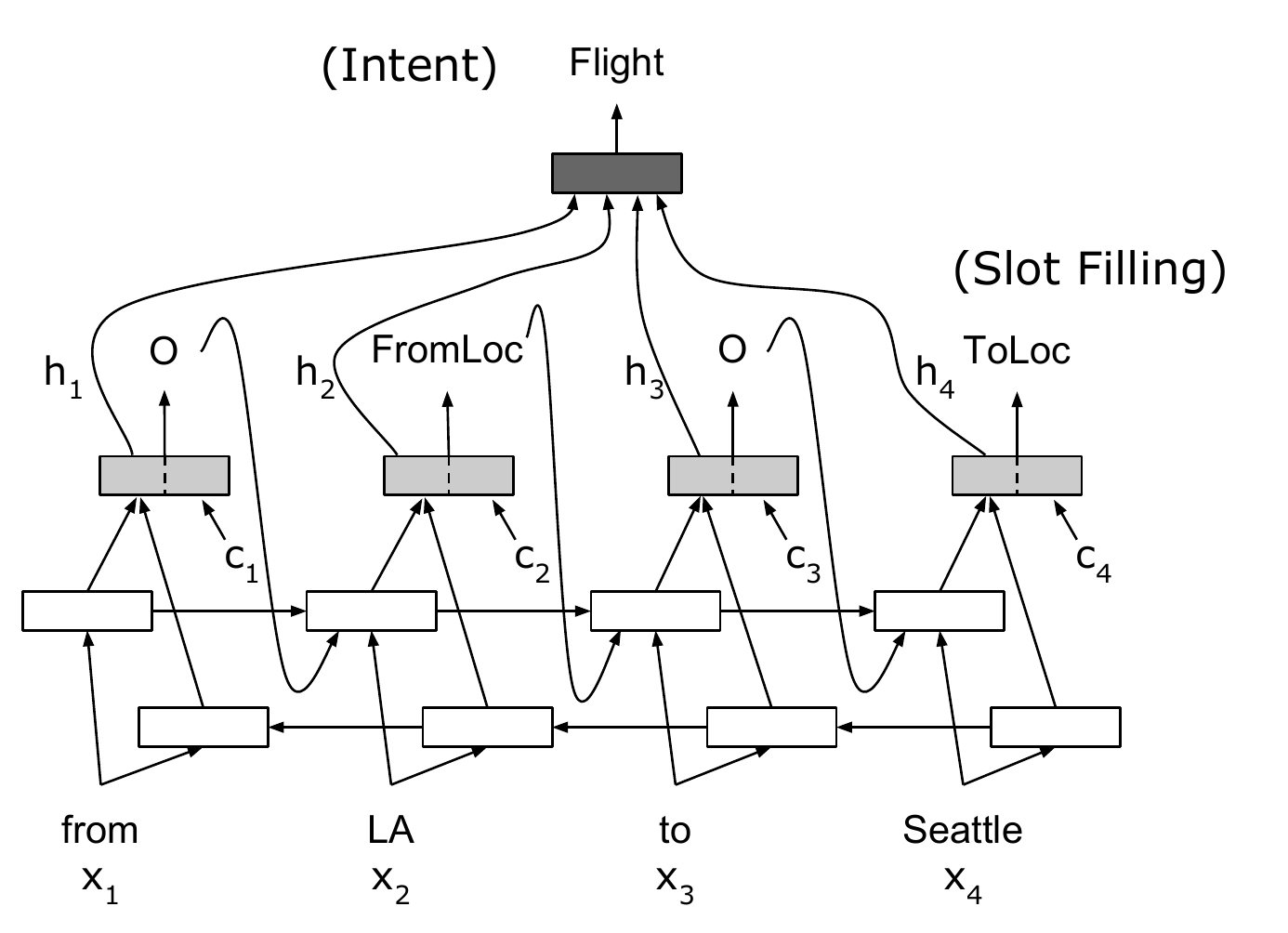}
            \caption{{\it Attention-based RNN model for joint intent detection and slot filling. The bidirectional RNN reads the source sequence forward and backward. Slot label dependency is modeled in the forward RNN. At each time step, the concatenated forward and backward hidden states is used to predict the slot label. If attention is enabled, the context vector $c_{i}$ provides information from parts of the input sequence that is used together with the time aligned hidden state $h_{i}$ for slot label prediction.}}
            \label{fig:attention_rnn}
        \end{figure}
        
        The attention-based RNN model for joint intent detection and slot filling is illustrated in Figure \ref{fig:attention_rnn}. The idea of introducing attention to the alignment-based RNN sequence labeling model is motivated by the use of attention mechanism in encoder-decoder models. In bidirectional RNN for sequence labeling, the hidden state at each time step carries information of the whole sequence, but information may gradually lose along the forward and backward propagation. Thus, when making slot label prediction, instead of only utilizing the aligned hidden state $h_{i}$ at each step, we would like to see whether the use of context vector $c_{i}$ gives us any additional supporting information, especially those require longer term dependencies that is not being fully captured by the hidden state.
        
        In the proposed model, a bidirectional RNN (BiRNN) reads the source sequence in both forward and backward directions. We use LSTM cell for the basic RNN unit. Slot label dependencies are modeled in the forward RNN. Similar to the encoder module in the above described encoder-decoder architecture, the hidden state $h_{i}$ at each step is a concatenation of the forward state $fh_{i}$ and backward state $bh_{i}$, $h_{i} = [fh_{i}, bh_{i}]$. Each hidden state $h_{i}$ contains information of the whole input word sequence, with strong focus on the parts surrounding the word at step $i$. This hidden state $h_{i}$ is then combined with the context vector $c_{i}$ to produce the label distribution, where the context vector $c_{i}$ is calculated as a weighted average of the RNN hidden states $\mathbf{h} = (h_{1}, ..., h_{T})$.
    
        For joint modeling of intent detection and slot filling, we reuse the pre-computed hidden states $\mathbf{h}$ of the bidirectional RNN to produce intent class distribution. If attention is not used, we apply mean-pooling \cite{zhang2015character} over time on the hidden states $\mathbf{h}$ followed by logistic regression to perform the intent classification. If attention is enabled, we instead take the weighted average of the hidden states $\mathbf{h}$ over time.
        
        Comparing to the attention-based encoder-decoder model that utilizes explicit aligned inputs, the attention-based RNN model is more computational efficient. During model training, the encoder-decoder slot filling model reads through the input sequence twice, while the attention-based RNN model reads through the input sequence only once.

\section{Experiments}

    \subsection{Data}
    ATIS (Airline Travel Information Systems) data set \cite{hemphill1990atis} is widely used in SLU research. The data set contains audio recordings of people making flight reservations. In this work, we follow the ATIS corpus\footnote{We thank Gokhan Tur and Puyang Xu for sharing the ATIS data set.} setup used in \cite{mesnil2015using, liu2015recurrent, xu2013convolutional, tur2010left}. The training set contains 4978 utterances from the ATIS-2 and ATIS-3 corpora, and the test set contains 893 utterances from the ATIS-3 NOV93 and DEC94 data sets. There are in total 127 distinct slot labels and 18 different intent types. We evaluate the system performance on slot filling using F1 score, and the performance on intent detection using classification error rate. 
    
    We obtained another ATIS text corpus that was used in \cite{xu2013convolutional} and \cite{jeong2008triangular} for SLU evaluation. This corpus contains 5138 utterances with both intent and slot labels annotated. In total there are 110 different slot labels and 21 intent types. We use the same 10-fold cross validation setup as in \cite{xu2013convolutional} and \cite{jeong2008triangular}.

    \subsection{Training Procedure}
    LSTM cell is used as the basic RNN unit in the experiments. Our LSTM implementation follows the design in \cite{zaremba2014recurrent}. Given the size the data set, we set the number of units in LSTM cell as 128. The default forget gate bias is set to 1 \cite{jozefowicz2015empirical}. We use only one layer of LSTM in the proposed models, and deeper models by stacking the LSTM layers are to be explored in future work. 
    
    Word embeddings of size 128 are randomly initialized and fine-tuned during mini-batch training with batch size of 16. Dropout rate 0.5 is applied to the non-recurrent connections \cite{zaremba2014recurrent} during model training for regularization. Maximum norm for gradient clipping is set to 5. We use Adam optimization method following the suggested parameter setup in \cite{kingma2014adam}. 
     
    \subsection{Independent Training Model Results: Slot Filling}

        We first report the results on our independent task training models. Table 1 shows the slot filling F1 scores using our proposed architectures. Table 2 compares our proposed model performance on slot filling to previously reported results.

        \begin{table} [h]
        \caption{\label{table1} {\it Independent training model results on ATIS slot filling.}}
        \vspace{2mm}
        \centerline{
        \begin{tabular}{l c c}
        \hline
        \textbf{Model} & \textbf{F1 Score} & \textbf{Average}\\
        \hline  \hline
        (a) Encoder-decoder NN & 81.64 & $79.66 \pm 1.59$  \\ with no aligned inputs \\ \hline
        (b) Encoder-decoder NN & 95.72 & $95.38 \pm 0.18$ \\ with aligned inputs \\ \hline
        (c) Encoder-decoder NN & \textbf{95.78} & $95.47 \pm 0.22$ \\ with aligned inputs \& attention \\ \hline
        \hline
        BiRNN no attention & 95.71 & $95.37 \pm 0.19$ \\ \hline
        BiRNN with attention & \textbf{95.75} & $95.42 \pm 0.18$ \\
        \hline
        \end{tabular}
        }
        \end{table}

\begin{figure}[h]
            \centering
            \includegraphics[width=\linewidth]{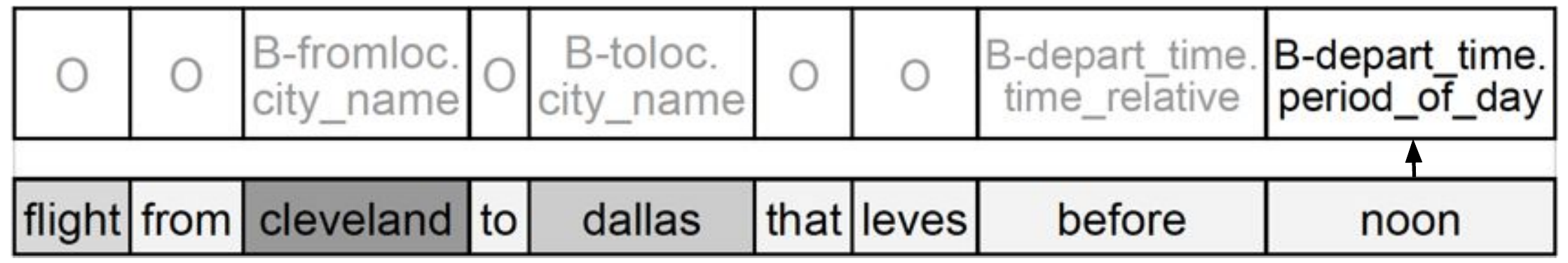}
            \caption{{\it Illustration of the inferred attention when predicting the slot label for the last word ``noon'' in the given sentence. Darker shades indicate higher attention weights. When word ``noon'' is fed to the model as the aligned input, the attention mechanism tries to find other supporting information from  the input word sequence for the slot label prediction. }}
            \label{fig:slot_weight_visualization}
        \end{figure}

        In Table 1, the first set of results are for variations of encoder-decoder models described in section 3.1. Not to our surprise, the pure attention-based slot filling model that does not utilize explicit alignment information performs poorly. Letting the model to learn the alignment from training data does not seem to be appropriate for slot filling task. Line 2 and line 3 show the F1 scores of the non-attention and attention-based encode-decoder models that utilize the aligned inputs. The attention-based model gives slightly better F1 score than the non-attention-based one, on both the average and best scores. By investigating the attention learned by the model, we find that the attention weights are more likely to be evenly distributed across words in the source sequence. There are a few cases where we observe insightful attention (Figure \ref{fig:slot_weight_visualization}) that the decoder pays to the input sequence, and that might partly explain the observed performance gain when attention is enabled.

        The second set of results in Table 1 are for bidirectional RNN models described in section 3.2. Similar to the previous set of results, we observe slightly improved F1 score on the model that uses attentions. The contribution from the context vector for slot filling is not very obvious. It seems that for sequence length at such level (average sentence length is 11 for this ATIS corpus), the hidden state $h_{i}$ that produced by the bidirectional RNN is capable of encoding most of the information that is needed to make the slot label prediction. 
        
        Table 2 compares our slot filling models to previous approaches. Results from both of our model architectures advance the best F1 scores reported previously.

        \begin{table} [h]
        \caption{\label{table2} {\it Comparison to previous approaches. Independent training model results on ATIS slot filling.}}
        \vspace{2mm}
        \centerline{
        \begin{tabular}{l c}
        \hline
        \textbf{Model} & \textbf{F1 Score}\\
        \hline  \hline
        CNN-CRF \cite{xu2013convolutional} & 94.35  \\ 
        RNN with Label Sampling \cite{liu2015recurrent} & 94.89\\ 
        Hybrid RNN \cite{mesnil2015using} & 95.06\\ 
        Deep LSTM \cite{yao2014spoken} & 95.08\\ 
        RNN-EM \cite{peng2015recurrent} & 95.25\\
        Encoder-labeler Deep LSTM \cite{kurata2016leveraging} & 95.66\\ 
        \hline
        Attention Encoder-Decoder NN  & \textbf{95.78} \\ (with aligned inputs) \\ 
        Attention BiRNN & \textbf{95.75} \\ 
        \hline
        \end{tabular}}
        \end{table}

    \subsection{Independent Training Model Results: Intent Detection}
    
        Table 3 compares intent classification error rate between our intent models and previous approaches. Intent error rate of our proposed models outperform the state-of-the-art results by a large margin. The attention-based encoder-decoder intent model advances the bidirectional RNN model. This might be attributed to the sequence level information passed from the encoder and additional layer of non-linearity in the decoder RNN. 
    
        \begin{table} [h]
        \caption{\label{table3} {\it Comparison to previous approaches. Independent training model results on ATIS intent detection.}}
        \vspace{2mm}
        \centerline{
        \begin{tabular}{l c}
        \hline
        \textbf{Model} & \textbf{Error (\%)}\\
        \hline  \hline
        Recursive NN \cite{guo2014joint} & 4.60  \\ 
        Boosting \cite{tur2010left} & 4.38  \\ 
        Boosting + Simplified sentences \cite{tur2011sentence} & 3.02 \\
        \hline
        Attention Encoder-Decoder NN  & \textbf{2.02} \\
        Attention BiRNN & \textbf{2.35} \\ 
        \hline
        \end{tabular}}
        \end{table}

    \subsection{Joint Model Results}
    
        Table 4 shows our joint training model performance on intent detection and slot filling comparing to previous reported results. As shown in this table, the joint training model using encoder-decoder architecture achieves 0.09\% absolute gain on slot filling and 0.45\% absolute gain (22.2\% relative improvement) on intent detection over the independent training model. For the attention-based bidirectional RNN architecture, the join training model achieves 0.23\% absolute gain on slot filling and 0.56\% absolute gain (23.8\% relative improvement) on intent detection over the independent training models. The attention-based RNN model seems to benefit more from the joint training. Results from both of our joint training approaches outperform the best reported joint modeling results.
    
        \begin{table} [h]
        \caption{\label{table4} {\it Comparison to previous approaches. Joint training model results on ATIS slot filling and intent detection.}}
        \vspace{2mm}
        \centerline{
        \begin{tabular}{l c c}
        \hline
        \textbf{Model} & \textbf{F1 Score}  & \textbf{Intent Error (\%)}\\
        \hline  \hline
        RecNN \cite{guo2014joint} & 93.22 & 4.60  \\ 
        RecNN+Viterbi \cite{guo2014joint} & 93.96 & 4.60  \\ 
        \hline
        Attention Encoder-Decoder  &  \textbf{95.87} & \textbf{1.57} \\ NN (with aligned inputs)\\ 
        Attention BiRNN & \textbf{95.98} & \textbf{1.79} \\ 
        \hline
        \end{tabular}}
        \end{table}
        
        To further verify the performance of our joint training models, we apply the proposed models on the additional ATIS data set and evaluate them with 10-fold cross validation same as in \cite{xu2013convolutional} and \cite{jeong2008triangular}. Both the encoder-decoder and attention-based RNN methods achieve promising results.

        \begin{table} [h]
        \caption{\label{table5} {\it Joint training model results on the additional ATIS corpus using 10-fold cross validation.}}
        \vspace{2mm}
        \centerline{
        \begin{tabular}{l c c}
        \hline
        \textbf{Model} & \textbf{F1 Score}  & \textbf{Intent Error (\%)}\\
        \hline  \hline
        TriCRF \cite{jeong2008triangular} & 94.42 & 6.93 \\ 
        CNN TriCRF \cite{xu2013convolutional} & 95.42 & 5.91 \\ 
        \hline
        Attention Encoder-Decoder  &  \textbf{95.62} & \textbf{5.86} \\ NN (with aligned inputs)\\ 
        Attention BiRNN & \textbf{95.78} & \textbf{5.60} \\ 
        \hline
        \end{tabular}}
        \end{table}

\section{Conclusions}
    In this paper, we explored strategies in utilizing explicit alignment information in the attention-based encoder-decoder neural network models.  We further proposed an attention-based bidirectional RNN model for joint intent detection and slot filling. Using a joint model for the two SLU tasks simplifies the dialog system, as only one model needs to be trained and deployed. Our independent training models achieved state-of-the-art performance for both intent detection and slot filling on the benchmark ATIS task. The proposed joint training models improved the intent detection accuracy and slot filling F1 score further over the independent training models.

  \newpage
  \eightpt
  \bibliographystyle{IEEEtran}

  \bibliography{mybib}


\end{document}